\documentclass[11pt,a4paper]{article}
\usepackage[hyperref]{acl2021}
\usepackage{times}
\usepackage{latexsym}
\usepackage{bm}
\usepackage{url}
\usepackage{tikz}
\usepackage{booktabs}
\usepackage{siunitx}
\usepackage{array}
\usepackage{multirow}
\usepackage{makecell}
\usepackage{amsmath}
\usepackage{cleveref}
\usepackage{footmisc}
\usepackage{verbatim}
\usepackage{amsfonts}
\usepackage{graphicx}
\usepackage{color}
\usepackage{float}
\usepackage{arydshln}
\usepackage{microtype}

\aclfinalcopy % Uncomment this line for the final submission
\def\aclpaperid{1602} %  Enter the acl Paper ID here

\title{Trade the Event: {C}orporate Events Detection for News-Based Event-Driven Trading}

\author{Zhihan Zhou \qquad  Liqian Ma \qquad  Han Liu \\
  Department of Computer Science, Northwestern University \\
  \tt \{zhihanzhou, liqianma\}@u.northwestern.edu\\ \tt hanliu@northwestern.edu }
\date{}

\begin{document}
\maketitle
\begin{abstract} 
% Existing news-based stock prediction models often utilize textual features (e.g., bag-of-words) and news sentiments to directly predict the stock movements, which may result in poor explainability and low signal-to-noise ratios.
In this paper, we introduce an event-driven trading strategy that predicts stock movements by detecting corporate events from news articles.
Unlike existing models that utilize textual features (e.g., bag-of-words) and sentiments to directly make stock predictions, we consider corporate events as the driving force behind stock movements and aim to profit from the temporary stock mispricing that may occur when corporate events take place.
The core of the proposed strategy is a bi-level event detection model. The low-level event detector identifies events' existences from each token, while the high-level event detector incorporates the entire article's representation and the low-level detected results to discover events at the article-level.  
We also develop an elaborately-annotated dataset \textbf{\emph{EDT}} for corporate event detection and news-based stock prediction benchmark. EDT includes $9721$ news articles with token-level event labels as well as $303893$ news articles with minute-level timestamps and comprehensive stock price labels.  
Experiments on EDT indicate that the proposed strategy outperforms all the baselines in winning rate, excess returns over the market, and the average return on each transaction. \footnote{Code and the EDT dataset is available at \url{https://github.com/Zhihan1996/TradeTheEvent}}

% forces the model to develop fine-grained understandings of each event, while 
% for corporate event detection and news-based intraday stock trading benchmark. 

% We propose an event detection model that not only detects the corporate events but also indicates where the detected event occurs in the original text. A joint optimization method is introduced to train the model to perform event detection on both article-level and token-level. 

% In contrast, we consider the corporate event as a more objective and convincing trading signal (e.g., \textit{positive sentiment} v.s. \textit{incoming stock split}). 

\end{abstract}

\section{Introduction}
By shaping investors' perceptions and assessments of companies, financial news has significant impacts on the stock market \citep{engle1993measuring, tetlock2007giving}. News-based stock prediction models are thus developed to automatically discover signals of stock market movements from the countless news articles that generated every moment. \citep{kalyani2016stock, shah2018predicting, 8848203}.

% \vspace{-8mm}

Previous studies mainly rely on textual features and sentiment analysis to forecast the stock movements \citep{fung2003stock, liu2018leveraging, 3155133}. Both of them, however, often face the problem of poor explainability and low signal-to-noise ratio. Textual feature-based methods often formulate the stock prediction as a text classification problem by directly predicting the rise and fall of stocks based on the extracted features. These models fail to make reasonable trading decisions since they omit the reasons behind stock price changes. Sentiment-based methods avoid this problem by regarding the news articles' sentiments as the indicator of stock movement. However, news sentiment is subjective, which can be greatly affected by the author's standpoint and writing style.

\begin{figure}[t]
		\centering
		\begin{tikzpicture}
		\draw (0,0 ) node[inner sep=0] {\includegraphics[width=1\columnwidth, trim={2.4cm 0.85cm 1.9cm 1.6cm}, clip]{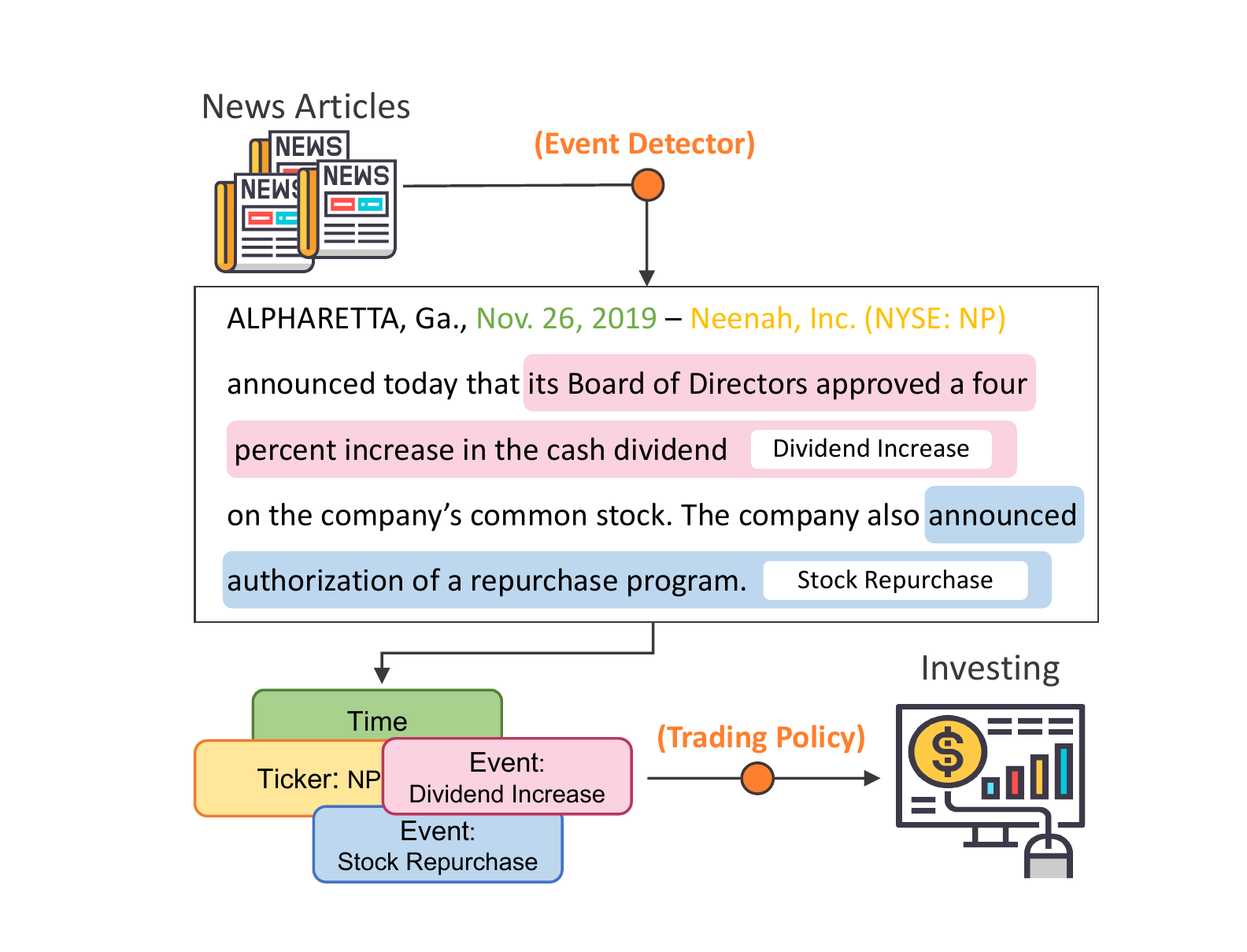}};
		\end{tikzpicture}
		\caption{Overview of the event-driven trading strategy}
		\label{fig:overview}

\end{figure}

% Event-driven strategies \citep{leinweber2011event, ding2015deep} consider events as the driving force behind stock price movements and aim to profit from the temporary stock mispricing, which may occur when corporate events take place. The core of an automated event-trading strategy is event detection.

Unlike textual features and sentiments, corporate events are objective facts that impact how investors perceive and assess the related companies. Thus, we resort to corporate events to make more convincing and explainable stock predictions.  \citet{baseline} achieves corporate events detection by splitting a news article into sentences and detecting events on each of them with multi-label sentence classification. This method, however, discards the global contextual information of the entire article and fails to indicate the evidence of events' existences. We believe that detecting events at a smaller granularity (e.g., at the token-level) is beneficial on both model training and application sides. During training, explicitly assigning a label to each token gives the model specific guidelines of what to identify. On the application side, each detected event is supported by one or more subsequences of the original article, allowing users to easily distinguish the predicted results.

However, singly detecting events from the token-level may result in a lack of macro understandings of the entire article. To tackle this, we introduce a bi-level event detection model, in which a low-level detector identifies the subsequences that describe specific events by classifying each token. And a high-level detector takes the predicted results from low-level and integrates them with the input article's global contextual information to predict the probabilities of each event's existence.

% We also perform a domain adaptation to equip the model with prerequisite domain knowledge. 

Another problem with existing models is that they ignore the timeliness of news articles. Most of them utilize news articles to predict the related securities' rise/fall on the following trading day(s). However, stock prices are very likely to change immediately in response to noteworthy news. Thus, the stock movement in the following trading day(s) may not accurately reflect a news article's influence. 
To tackle this, we make stock predictions as soon as a news article is published and perform tradings at that moment with the proposed trading policies.

Based on the event detection model and trading policies, we construct an event-driven trading strategy that simultaneously detects corporate events from news articles, indicates the subsequences that describe the detected events, and performs trading on the related stocks. 
% We believe computers have an advantage when trading on unanticipated events since their superiority in speed and scalability enables them to extract useful information from a massive amount of sources. 
By running the strategies against massive historical data in EDT, we demonstrate the superiority of the proposed strategy in terms of excess returns over the market and the average return on each transaction. The experiment results also reveal the timeliness of news and the effectiveness of corporate events in indicating stock movement.

% Beside, cases studies indicates that the corporate event is more interpretable and reliable than news sentiments.

The main contributions of our paper are as summarized follows: 
(i) We introduce a novel event-driven trading strategy that detects trading signals from arbitrary unlabeled news articles; 
(ii) We present \textbf{\emph{EDT}}, an elaborately-annotated dataset with $300000$+ news articles for corporate event detection and news-based trading benchmark.
(iii) We propose a bi-level event detection model that integrates macro and fine-grained understandings to effectively identify corporate events;

% \vspace{-2mm}

\section{Problem Definition}

% \vspace{-2mm}

\label{section:problem_definition}
We aim to construct an event-driven trading strategy that automatically detects corporate events from news articles and performs trading accordingly. The proposed strategy consists of two components: \textit{bi-level event detector} and \textit{trading policy}.

The low-level event detector identifies events from each token. We formulate the low-level event detection as a sequence labeling problem. The label set $\mathcal{L} = \left\{e_1,e_2,...,e_k,O\right\}$ consists of $k$ pre-defined events and a special label $O$ that stands for \textbf{\emph{Noevent}} \footnote{events that are not included in $\left\{e_i\right\}_{i=1}^{k}$ are also considered as \emph{Noevent}}. We define an article $\bm{x} = (x_1,x_2,...,x_n)$ as a sequence of tokens and define its label sequence as $\bm{y} = (y_1,y_2,...,y_n)$. The same event may be mentioned multiple times in an article, and a single article may contains multiple events. If a subsequence $\bm{x'} = (x_t,x_{t+1},...,x_{t+s})$ of $\bm{x}$ describes the \textbf{\emph{event i}}, $\left\{y_j\right\}_{j=t}^{t+s}$ are labeled as $e_i$. All the other words are labeled as $O$. The low-level event detection is defined as follows: given an article $\bm{x^*} = (x_1,x_2,...,x_n)$, find its best label sequence $\bm{y^*} = (y_1, y_2, . . . , y_n)$. 
We say \textbf{\emph{event i}} is detected by the low-level detector if $e_i \in \bm{y^*}$. 
Based on the low-level detected results, the high-level event detector calculates the probability of each event's existence. We say \textbf{\emph{event i}} is detected by the high-level detector if its existence probability is larger than a given threshold. We combine the predictions on both levels as the final prediction.
When events are detected, the \textit{trading policy} decides when to buy and sell the related securities.

\section{Strategy}
This section discusses the proposed strategy by respectively introducing the event detector and the trading policies. 

\subsection{Event Detector}
Before training the model to detect events, we first equip it with prerequisite knowledge of the financial domain by performing a \textit{domain adaptation}.

% \vspace{-5mm}

\subsubsection{Domain Adaptation}

Since the same event can be expressed with significantly different terms and descriptions, and the same terms can refer to different meanings, understanding the event itself and its related terms is of great importance. 
% \textit{Guidance} is a good example here. In general, this word is closely related to \textit{advice} or \textit{instruction}, while in corporate finance, it refers to a company's estimate of its upcoming-quarter/fiscal year earnings. Thus, if it is expressed as \textit{estimation of Q3 earning}, models trained on general domain texts may fail to detect it. 
We perform domain adaptation by training the model with Masked-Language-Model (MLM) loss on a financial encyclopedia as well as financial news articles. Section \ref{section:data_da} discusses this corpus in details. During training, 15\% tokens of an input sequence are masked, and the model is asked to predict the masked tokens. The prediction is essentially a multi-class classification over the entire vocabulary. We optimize the model with the \textit{Categorical CrossEntropy} loss calculated on all the masked tokens.

% \begin{figure}[h]
% 		\centering
% 		\begin{tikzpicture}
% 		\draw (0,0 ) node[inner sep=0] {\includegraphics[width=1\columnwidth, trim={1.9cm 4cm 2.15cm 4.75cm}, clip]{./picture/2.pdf}};
% 		\end{tikzpicture}
% 		\caption{An example of how the corporate event “guidance” can be expressed in other financial terms, and its other usages that may cause ambiguity.}
% 		\label{fig:guidance}
% \end{figure}

\subsubsection{Bi-Level Event Detection}

\begin{figure}[t]
		\centering
		\begin{tikzpicture}
		\draw (0,0 ) node[inner sep=0] {\includegraphics[width=1\columnwidth, trim={3.5cm 1.85cm 3.3cm 2.1cm}, clip]{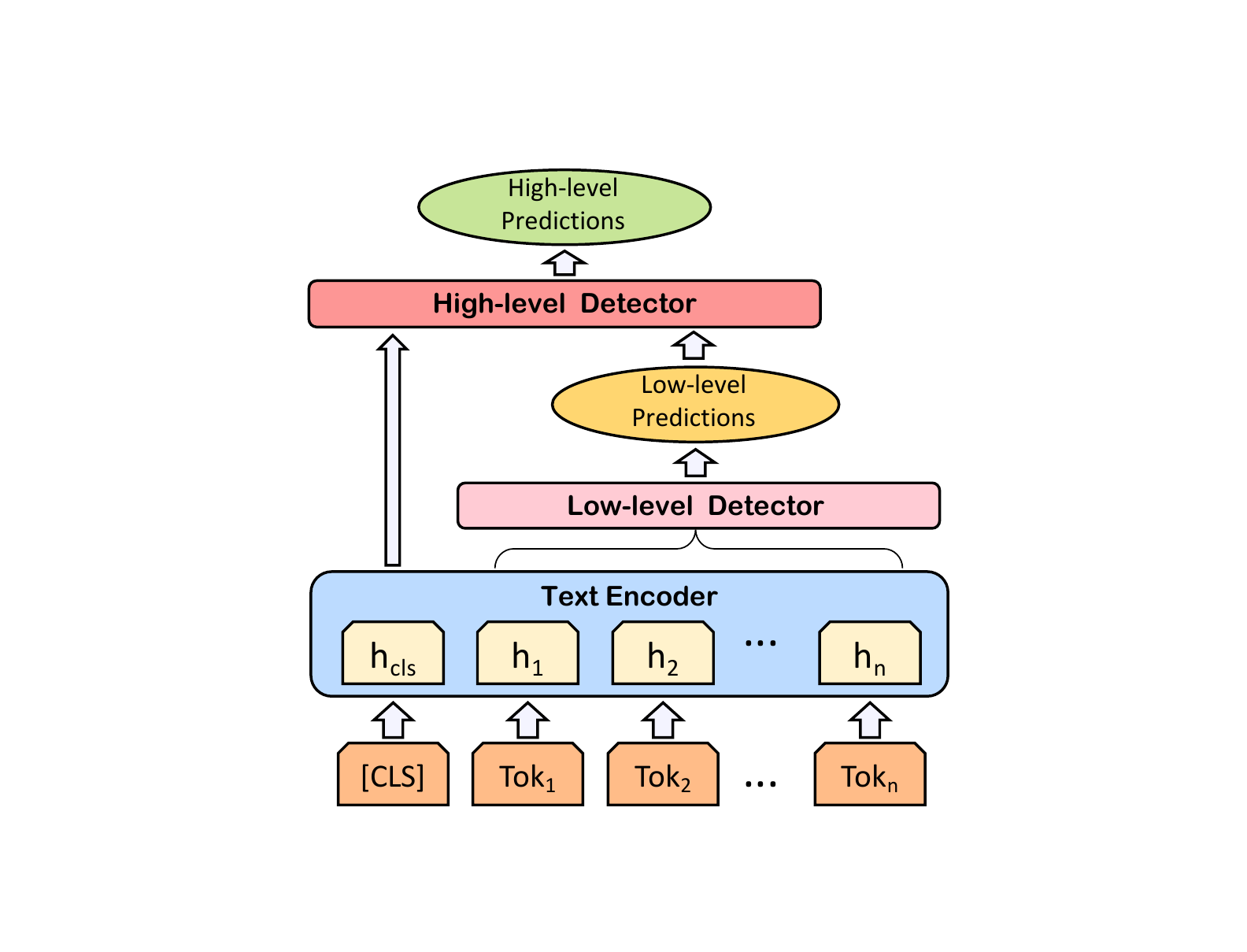}};
		\end{tikzpicture}
		\caption{Model overview. The low-level detector identifies corporate events from each token, while the high-level detector summarizes the low-level predictions and the input representation to detect events at the article-level}
		\label{fig:model}
\end{figure}

As shown in \cref{fig:model}, the event detection model takes an article as input and respectively detects events from two levels. Each article is concatenated with a special token [CLS]. The Transformer-based text encoder calculates a series of hidden states for each token. We consider [CLS]'s last hidden state $h_{cls}$ as the representation of the entire article, and $h_i$ as the representations of \emph{token i}. 

The low-level detector identifies the subsequences that describe corporate events. For a given label set  $\mathcal{L}$ (\cref{section:problem_definition}) with $K+1$ labels (e.g., $K$ pre-defined events and a ``\emph{Noevent}''), the detector assigns $K+1$ scores to each token by performing a multi-classes classification based on the token's representation. 
Each score stands for the confidence of finding a specific event at this position. These scores are concatenated and passed to the high-level detector.
During training, we calculate the \textit{Categorical CrossEntropy} loss for each token of an article and average them as the article's low-level loss. 
% Let $\hat{y_i}$ and $y_i$ as the prediction results and label for \textbf{\emph{token i}}, we calculate the token-level loss for an article of length $N$ as:
% $L_1 = \frac{\sum_{i=1}^{N} CrossEntropy(y_i, \hat{y_i})}{N}$.

The high-level detector concatenates the low-level prediction as well as the entire article's representation to calculate the probability of each event's existence. We formulate it as a multi-label classification problem. Specifically, we assign $K$ binary labels to each article and represent them with a $K$-length label vector. For articles without any events, their label vectors are all-zero. If \emph{event i} occurs in an article, the $i$-th component of its label vector is set to 1.
We utilize the \textit{Binary CrossEntropy with Sigmoid} as the loss function. This function considers the $K$-label classification as $K$ independent binary classification problems. Specifically, it uses the $Sigmoid$ function to map each vector component of the high-level detector's output to $(0,1)$.
We consider the mapped score of each event as its probability of existing in the input article.
We then calculate the \textit{Binary CrossEntropy} loss between the mapped score and the binary label of each event. The losses of all the events are summed as the high-level detector's loss. 
We sum the losses of the low-level and high-level detectors to simultaneously optimize the detectors and the text encoder.

\subsubsection{Ticker Recognizer}
To make the proposed strategy applicable to arbitrary news articles, besides detecting events, we also recognize the related securities (e.g., company) to trade on. 

Each security listed on an exchange has a unique ticker, which is a unique arrangement of characters (e.g., Amazon's ticker at the NASDAQ exchange is AMZN). 
% When events are detected, we recognize the related company and its ticker from the news article to trade. 
To recognize the tickers, we download company-ticker pairs (e.g., Amazon v.s. AMZN) for all the securities listed on NYSE and NASDAQ from Yahoo\footnote{\url{finance.yahoo.com}}. For a given article, we perform string matching between the article and all the company-ticker pairs. If multiple company-ticker pairs are matched, we choose the one that occurs the most times. Some tricks are employed to improve the accuracy and efficiency. For example, the company-ticker pairs that match the title's first few words are assigned higher confidence. Although a single article may include multiple securities, we only recognize the one that occurs most to simplify the setting.

\subsection{Trading Policy}
\label{section:policy}
To minimize other factors' influences, in this paper, we trade only on stocks instead of any derivatives (e.g., options). We relate each detected event to a \textit{long} or \textit{short} trading signal singly based on its event type. Events that may result in a stock price rise are considered as \textit{long} signals, while events that may lead to a fall of stock prices are considered as \textit{short} signals.

We implement two trading policies named \textbf{Trade-At-End} and \textbf{Trade-At-Best}. Both of them long (e.g., buy) the related stocks for \textit{long} signals and perform short-selling for \textit{short} signals. We define a transaction as a buy(sell) and a sell(buy) of the same stock. The policies always start a transaction (e.g., perform a buy or a short-selling) at the first available time when an event is detected. 
% For events that are detected outside the market hours, the policies trade the stock at the open time of the next trading day. We take buying as a example.

\textbf{Trade-At-End ($k$):} This policy holds a started transaction for $k$ trading days and closes the transaction on the last day when the market closes. 
% This policy estimates the profit that a trader can gain at least within $k$ trading days with a detected event.
% It sells stocks bought before 13:00 ET on the same day and sells the other stocks on the next trading day. It estimates the profit that an unskilled day trader can gain at least with a detected event.

\textbf{Trade-At-Best ($k$):} This policy closes a transaction at the best price (e.g., highest for sell and lowest for buy) within $k$ trading days from the start date. It estimates the profit that a trader can gain at most within $k$ trading days with a detected event.
% This policy sells the bought stocks at the highest price before the market closes. Although it is impossible to implemented this policy in the real-world, it estimates the profit that a trader can gain at most in the same day the event is detected.

\section{Data}
\label{sec:data}
In this section, we discuss the \textit{EDT} dataset. EDT contains data for three purposes: 1. corporate event detection (\cref{section:data_et}); 2. news-based trading strategy benchmark (\cref{section:data_e}); 3. financial domain adaptation (\cref{section:data_da}). 

% We start from the event type (\cref{section:data_et}), and then discuss the data for event detection model training/validating (\cref{section:data_e}) and the data for news-based trading strategy evaluation (\cref{section:data_e}). We also introduce the data used for domain adaptation in \cref{section:data_da}.

% the events as well as data collection, pre-processing, and annotation. We start from brief introductions of the events and follow with discussions of data for training (\cref{section:data_sl}), domain adaptation (\cref{section:data_da}), and evaluation (\cref{section:data_e}). 

\subsection{Data for Event Detection}
\label{section:data_et}
% We first presents the pre-defined events and then introduce the labeled data in details.
% In order to make convincing and reasonable trading decisions, instead of investigating general events, 
We choose 11 types of corporate events that have relatively predictable and straightforward impacts on the stock price based on financial knowledge.

\subsubsection{Event Type} 

In this work, we only focus on non-periodic corporate events. Trading on periodic corporate events such as \textit{Earning Call} is much trickier since investors can access their information from multiple sources in advance. We leave them for future works.

\textbf{Guidance Increase (GI)}
Guidance is a company's public estimates of its upcoming-quarter/fiscal year earnings. This event includes the announcements of guidance increase or upgrade.

\textbf{Acquisition (A)}
An acquisition event happens when a company announces to purchase all or a portion of another company's shares or assets.

\textbf{New Contract (NC)}
The new contract event refers to a company announcement of being awarded a new contract.

\textbf{Stock Split (SS)}
A stock split event refers to a company that divides the existing shares of its stock into multiple new shares.

\textbf{Reverse Stock Split (RSS)}
This is the reverse process of the stock split, which consolidates the number of existing stock shares into fewer shares.

\textbf{Positive Clinical Trial \& FDA Approval (CT)}
This event includes (i) positive trial results from clinical studies; (ii) receiving FDA approval, clearance, or being granted by FDA to market legally in the United States.

\textbf{Stock Repurchase (SR)}
A company's stock repurchase events include declaring, reinstatement, or increasing a stock buyback plan. 

\textbf{Dividend (RD)}
The dividend is a distribution of some of a company's profits paid to its shareholders. 

\textbf{Dividend Cut (DC)}
A dividend cut means to reduce, stop, or suspend a pre-announced dividend. 
% Suspension of a stock repurchase program is also labeled as this event.

\textbf{Dividend Increase (DI)}
A dividend increase refers to an increase in the regular dividend.

\textbf{Special Dividend (SD)}
A special dividend is an event that a company declares a non-recurring dividend paid to its shareholders.

\subsubsection{Detailed information}
\label{section:data_sl}

\begin{table}[h]
    \centering
	\footnotesize
	\setlength{\tabcolsep}{3mm}{
	\begin{tabular}{lrlr}\toprule
		{\textbf{Event}} & {\textbf{Num.}} 
		&{\textbf{Event}} & {\textbf{Num.}}\\

		\midrule

		{A } & 229 & {CT } & 314 \\
		{RD } & 290 & {DC } & 109  \\
		{DI } & 225 & {GI } & 99  \\
		{NC } & 518 & {RSS } & 51  \\
		{SD } & 80 & {SR } & 385  \\
		{SS } & 76 & \\
%         {A } & 229  & {DC } & 109  \\
% 		{DI } & 225 & {GI } & 99  \\
% 		{NC } & 518 & {RSS } & 51  \\
% 		{SD } & 80 & {SR } & 385  \\
% 		{SS } & 76 \\ 

		\bottomrule
	\end{tabular}}
	\caption{ \footnotesize  
		Number of articles in \textit{EDT} for each event.
	}\label{tb:data}
\end{table}

We collect 9721 English news articles, of which 2266 articles contain at least one of the above events. Table \ref{tb:data} shows the number of articles that corresponds to each event. The rest 7455 articles are news articles that do not contain any of the above events. We expect them to help the model better distinguish the event-related articles from the non-event ones. Among them, we deliberately include hundreds of non-event articles that are highly similar to event-related ones.
An example here is ``Apple announces a stock repurchase program'' v.s. ``Apple announces the completion of the recently announced stock repurchase program'', in which we do not expect the latter one to have a significant influence on the stock price. 
% We consider these articles as ``adversarial examples'' that may increase the robustness of models trained on this dataset. 

These news articles are downloaded from PRNewswire\footnote{\url{https://www.prnewswire.com/}}, Businesswire\footnote{\url{https://www.businesswire.com/}} and GlobeNewswire\footnote{\url{https://www.globenewswire.com/}} using keywords-search. The keywords for each event are manually determined based on samples of that event. Each article's title, subtitle, and main text are concatenated after data cleaning (e.g., remove special symbols). We annotate each article with token-level labels. Two human annotators are asked to independently mark the subsequences that best describe the pre-defined corporate events. The annotations of an article are produced if the annotators give the same result. Otherwise, they discuss the best annotations. 

We randomly sample 80\% articles of each event and combine them with 80\% of non-event articles to form the training data. The rest are considered as the validation data. 

% \vspace{-2mm}

% \vspace{-2mm}
\subsection{Data for Strategy Evaluation}
\label{section:data_e}                                                                                                                                        
% Two types of data are required for model evaluation: news articles with timestamps and minute-level stock prices. 

We develop this data to benchmark news-based stock prediction models and trading strategies. To accurately account for the stock movement, the news articles should be original-sourced. To mimic the real-world situation, the news articles should be diverse enough (e.g., in different categories). Thus, we choose PRNewswire and Businesswire as the article collection sources and download all the English news from PRNewswire (Mar 1st, 2020 - Apr 30th, 2021) and Businesswire (Aug 16th, 2020 - May 6th, 2021). We remove the articles that exist in the training data (\cref{section:data_sl}). Since some pre-defined events are infrequent (e.g., stock split), to ensure that there are at least a few samples of every event, we add all the articles of the validation data (\cref{section:data_sl}) to this data. After data cleaning, there are $303893$ news articles.

Each article comes with a minute-level timestamp, which allows researchers to locate the exact event happening time. 
Generally, news-based trading strategy evaluation involves four steps. (i) Identify trading signals (e.g., corporate events or sentiments) from news articles; (ii) For each article where trading signals are detected, recognize the related company(ticker); (iii) Get the recognized company's stock price data around the publish time of the news; (iv) Perform transactions based on trading policies. 

To enable researchers without ticker recognizers and historical stock price data to benchmark their models/strategies, we assign each article with an automatically recognized ticker as well as detailed price labels of that ticker. With the detailed price labels of each news article, strategy evaluation can be as easy as ``counting the price changes on the articles that are recognized as trading signals''.

Specifically, an article's price labels includes: \textit{open} / \textit{close} prices at the first minute we can trade on, \textit{highest} / \textit{lowest} prices in the following 1/2/3 trading days, \textit{close} prices in the following 1/2/3 trading days, and the minute-level timestamp corresponding to each price. If available, we take the prices in the pre-market and after-hours into consideration since many corporate events are announced in these periods, and stock prices may change greatly during these times. The price labels are empty for articles where no ticker is recognized or the historical price data is unavailable. Among all the evaluation data, $106619$ articles come with non-empty price labels.\footnote{We acquire historical stock price data from \url{https://polygon.io/}}
% Take the stock price change rate in 1 day as an example. For articles published during the market hours (e.g., Mon-Fri 9:30 am to 4 pm ET in the U.S), we calculate the change rate between the ticker's prices at the article's publish time (e.g., 11:32 am) and the market's close time (e.g., 4:00 pm). For other articles, we calculate the change rate between the ticker's open and close prices on the next trading day. 

% The timestamped news articles are used for trading signals (e.g., sentiments and corporate events) detection. Since the market reacts quickly to corporate events, 
% we regard the publishing time of a news article as the happening time of the events detected from it. 
% Thus, each news article should come from websites that provides first-hand news. 
% Moreover, to make the experiment environment similar to the complicated real stock environment, we need as many different news as possible emerged in a certain period of time. 
% We download all the English news from PRNewswire (Mar 1st - Sep 1st) and Businesswire (Aug 16th - Oct 10th) with minute-level timestamps. Since some pre-defined events are very rare (e.g., stock split), to ensure that there are at least a few samples of every event, we remove the labels and add all the articles of the validation data(\cref{section:data_sl}) to the timestamped news articles. 
% Minute-level stock price data is acquired from Polygon.io\footnote{https://polygon.io/}. 

\subsection{Data for Domain Adaptation}
\label{section:data_da}

The corpus for domain adaptation contains financial news articles and a financial terms encyclopedia, which is considered as unstructured domain knowledge. For encyclopedia, we download 6260 explanatory documents from Investopedia\footnote{\url{https://www.investopedia.com/}}. Each document explains a specific financial term and describes the role it may play in the financial market. For news articles, we directly utilize all the news articles of the training data (\cref{section:data_sl}).

\section{Experiment}

In this section, we first exhaustively compare Bi-level Detection with baselines under different settings. Then, we discuss how each event contributes to the overall trading results. Lastly, we analyze the profitability and practicality of the proposed strategy in real-world stock trading.  

Among the 11 corporate events in the EDT dataset, we do not trade on the \textbf{Dividend} since we do not consider it to have a significant influence on stock prices. Among the rest, we consider \textbf{Reverse Stock Split} and \textbf{Dividend Cut} to have negative influences on the stock price, while the others to have positive influences. We evaluate the performance of the proposed strategy with backtesting. Backtesting is widely used to evaluate a trading strategy's effectiveness by running the strategy against historical data. We perform trade on all the detected trading signals for each model.

\paragraph{Metrics} For a ``buy'' transaction, we define its return as $\frac{P_{sell}-P_{buy}}{P_{buy}}\%$, while for a ``short-selling'', we define its return as $\frac{P_{sell}-P_{buy}}{P_{sell}}\%$. Here, $P$ stands for the price. If a transaction's return is greater than or equal to 0, we call it a ``win''. If a transaction's return is greater than or equal to 1\%, we call it a ``big win''. For each model, we calculate its \textbf{winning rate}, \textbf{big win rate} (rate of big wins among all the transactions) and \textbf{average return on each transaction}.
We also evaluate the models' excess returns over the market, where we consider the \textit{S\&P 500} index as the benchmark of the market performance. The \textbf{market return} is estimated as the return of buying \textit{S\&P 500 index ETF} for $\$10000$ on Mar 1st, 2020 and sell all of them on May 6th, 2021\footnote{In accordance to the time span of data for evaluation}. For each model, we start with $\$10000$ cash and invest $\$2000$ to each trading signal. When available cash is less than $\$2000$, we invest 20\% of available cash to the detected signal. 
We report the \textbf{excess returns} of each model, which equals to a model's \textbf{total returns} minus the \textbf{market return}. \footnote{Since Trade-At-Best always finishes the transaction at the best price, its winning rate is always 100\% and its total returns is almost linearly related to the number of transactions. Thus, we only report the average return of this policy. } We assume there is a $0.3\%$ commission fee on each transaction.

% \paragraph{Backtesting} 
% For each trading policy, we conduct two lines of experiments. One starts transactions with the \textit{Open} price at the minute that the trading signal is identified (e.g., price at 11:23:00), while the other starts with the \textit{Close} price at that minute (e.g., price at 11:23:59)\footnote{All the price are adjusted}. 

% \paragraph{Trading Policies} We use ``TAE-\textit{Open} (1)'' to represent a Trade-At-End policy that starts transaction with the \textit{Open} price and hold the transaction for 1 day. Other names follow the same rule.

\paragraph{Model Hyperparameters} We employ the pre-trained BERT \citep{bert} model as the text encoder. Specifically, we use the \textit{bert-base-cased} checkpoint. Both the low-level and high-level detectors consist of a hidden layer and an output layer. There are 2048 hidden units in the hidden layer. We utilize AdamW optimizer with batch size 32 and learning rate 5e-5 to train the encoder and detectors together for 5 epochs. We set the maximum input length as 256 since we find almost all the events mentioned in a news article exist in its first 256 tokens. Training of the model costs 15 minutes on 4 Nvidia RTX 2080Ti GPUs. We conduct each experiment with 3 different random seeds and report the average results.

\paragraph{Trading Details} For Trade-At-End, we execute a stop loss of 20\% (e.g., sell a stock immediately when it falls 20\%). In all the experiments, we only trade on the news articles where the historical price data of the detected ticker is available at the minute when the article is published. In other words, we ignore all the news articles that are not published during the market hours and articles where the historical price data is incomplete.

% \zhihan{emphasize Derivative and only trade on highly-confident events may invites much more profit}

% \zhihan{accuracy for ticker recognition}

\subsection{Baseline}

\paragraph{Vader} \citep{gilbert2014vader} is a rule-based sentiment analysis model that assigns \textit{positive}, \textit{negative}, and \textit{neutral} scores to an article. We consider news articles with a positive score greater than $0.2$ as \textit{long} trading signals.

\paragraph{BERT-SST} is a BERT-based \citep{bert} sentiment analysis model trained on the Stanford sentiment treebank (SST) dataset. We respectively consider news articles with a positive score greater than $0.995$ and $0.9$ as \textit{long} trading signals to reduce the threshold's influences on the final results.

% \paragraph{Keywords} detects corporate events by keywords matching. For example, news articles contains \textit{special} and \textit{dividend} are considered to contain the \textit{Special Dividend} event. We consider each event as a trading signal. This method measures the average market movements resulting from the news articles that may contain corporate events.

\paragraph{Sentence} \citep{baseline} splits an article into sentences and performs sentence-level event detection based on multi-label text classification. It was original implemented with SVM and LSTM. We re-implement it with BERT to compare it fairly with our models. We split each article into sentences with the NLTK toolkit \citep{loper2002nltk} to train and evaluate the model.

\paragraph{BERT-CRF} \citep{CRF} was originally proposed as a Conditional Random Fields-based sequence labeling model, which combines emission scores given by BERT and learned transition scores to find the global optimal label sequence for each input. We re-implement it to perform event detection singly on the token-level. Following the literature, we use different learning rates for the CRF(1e-3) and the BERT(3e-5) components.

\begin{table*}[t]
	\centering
	\footnotesize
	\setlength{\tabcolsep}{2.4mm}{
	\begin{tabular}{lrrrrr}\toprule
		\multirow{3}{*}{\textbf{Model}} &
		\multicolumn{3}{c}{\textbf{TAE (1)}} & \multicolumn{1}{c}{\textbf{TAB (1)}} & \\
		\cmidrule(lr){2-4} \cmidrule(lr){5-5}
		
		& {\textbf{Win Rate}} & {\textbf{Ave. Return}} & {\textbf{Exc. Returns}}   & {\textbf{Ave. Return}}  & {\textbf{Num. Trans.}}
 \\

		\midrule
		{\textbf{Vader} \citep{gilbert2014vader}}  & 52.8$||$24.3\% & 0.06\% & -\$8116 & 1.72\% & 4327\\
		
		{\textbf{BERT-SST (0.995)} }   & 54.3$||$26.9\% & 0.45\% & \$3743 & 2.94\% & 2378 \\
		
		{\textbf{BERT-SST (0.9)} } & 52.9$||$26.1\% & 0.41\% & \$44049 & 3.31\% & 15663 \\

		{\textbf{Sentence} \citep{baseline}}  & \textbf{55.5}$||$30.9\% & 1.37\% & \$54064 & 7.21\% & 2881 \\
		
		{\textbf{BERT-CRF}}  & 53.7$||$33.8\% & 1.60\% & \$83120 & 8.87\% & 3533 \\
        
        \midrule
        
		{\textbf{Bi-level Detection} (ours)} & 54.5$||$\textbf{34.2}\% & \textbf{1.74}\% & \$\textbf{84443} & \textbf{9.11}\% & 3118  \\

		\bottomrule
	\end{tabular}}
	\caption{ \footnotesize  
		This table shows the 1-day trading result, in which we start each transaction at the news article's publish time and end the transaction after 1 trading day. \textbf{Win Rate} stands for the overall winning rate (rate of transactions that have a return over 0) $||$ big win rate (rate of transactions that have a return over 1\%). \textbf{Ave. Return} stands for the average return on each transaction. \textbf{Exc. Return} stands for the total excess returns over the market when starting with \$10000 and invest \$2000 to each detected trading signal. \textbf{Num. Trans.} stands for the number of transactions (valid trading signals) of each model.
	}\label{tb:allwin1open}
\end{table*}

\begin{table*}[t]
	\centering
	\footnotesize
	\setlength{\tabcolsep}{2.4mm}{
	\begin{tabular}{lrrrrr}\toprule
		\multirow{3}{*}{\textbf{Model}} &
		\multicolumn{3}{c}{\textbf{TAE (2)}} & \multicolumn{1}{c}{\textbf{TAB (2)}} & \\
		\cmidrule(lr){2-4} \cmidrule(lr){5-5}
		
		& {\textbf{Win Rate}} & {\textbf{Ave. Return}} & {\textbf{Exc. Returns}}   & {\textbf{Ave. Return}}  & {\textbf{Num. Trans.}}
 \\

		\midrule
		{\textbf{Vader} \citep{gilbert2014vader}}  & 53.7$||$36.9\% & 0.38\% & -\$2551 & 3.11\% & 4327 \\
		
		{\textbf{BERT-SST (0.995)} }  & \textbf{53.8}$||$38.4\% & 0.62\% & \$8479 & 4.72\% & 2378 \\
		
		{\textbf{BERT-SST (0.9)}}  & 52.3$||$37.2\% & 0.46\% & \$12802 & 3.93\% & 15663 \\

		{\textbf{Sentence} \citep{baseline}}  & 52.7$||$39.9\% & 1.24\% & \$24673 & 9.49\% & 2881 \\

		{\textbf{BERT-CRF}} & 51.3$||$39.9\% & 1.39\% & \$52891 & 11.27\% & 3533  \\
        
        \midrule

		{\textbf{Bi-level Detection} (ours)} & 52.3$||$\textbf{40.8}\% & \textbf{1.56}\% & \$\textbf{59375} & \textbf{11.53}\% & 3118  \\

		\bottomrule
	\end{tabular}}
	\caption{ \footnotesize  
		This table shows the 2-day trading result, in which we start each transaction at the news article's publish time and end the transaction after 2 trading day.
	}\label{tb:allwin2open}
\end{table*}

\begin{table*}[t]
	\centering
	\footnotesize
	\setlength{\tabcolsep}{2.6mm}{
	\begin{tabular}{lcrrrrrrrrrrr}\toprule
		\multirow{2}{*}{\textbf{Metric}} & \multirow{2}{*}{\textbf{Policy}} &
		\multicolumn{10}{c}{\textbf{Event Type}} & \\
		\cmidrule(lr){3-12}

		 &   & {\textbf{A}} & {\textbf{CT}} & {\textbf{DC}} & {\textbf{DI}} & {\textbf{GI}} & {\textbf{NC}} & {\textbf{RSS}} & {\textbf{SD}}
		& {\textbf{SR}} & {\textbf{SS}}  \\

		\midrule

        \multirow{2}{*}{\textbf{WR. \%}} &
		{TAE (1)} & 52.8 & 51.7 & 60.0 & 66.9 & 55.9 & 53.3 & 57.8 & 84.0 & 60.2 & 43.3 \\
        & {TAE (3)} & 53.5 & 48.0 & 49.4 & 60.2 & 52.8 & 50.5 & 55.6 & 64.0 & 55.6 & 81.1 \\
        \midrule
        \multirow{4}{*}{\textbf{AR. \%}} &
        {TAE (1)} & 2.15 & 1.89 & 0.75 & 0.33 & 0.88 & 1.72 & 3.26 & 4.76 & 1.27 & -0.04 \\
        &{TAE (3)} & 1.93 & 2.05 & 0.02 & 1.02 & 1.19 & 1.32 & 4.25 & 5.49 & 1.51 & 4.99 \\
        &{TAB (1)}  & 9.98 & 12.32 & 5.71 & 1.74 & 4.63 & 9.40 & 8.19 & 8.02 & 5.26 & 4.27 \\
        &{TAB (3)}  & 13.46 & 17.51 & 10.95 & 4.21 & 7.73 & 13.24 & 19.68 & 13.13 & 8.79 & 17.91 \\
        \midrule
        \textbf{Num.} & & 960 & 721 & 39 & 135 & 282 & 697  & 26 & 29 & 225 & 5 \\  

		\bottomrule
	\end{tabular}}
	\caption{ \footnotesize  
		The table shows \textbf{Bi-level Detection}'s 1-day and 3-day trading results on each event, where \textbf{WR.} stands for the winning rate and \textbf{AR.} stands for the average return on each transaction.
		Each column respectively shows the results of Acquisition(A), Clinical Trial(CT), Dividend Cut(DC), Dividend Increase(DI), Guidance Increase(GI), New Contract(NC), Reverse Stock Split(RSS), Special Dividend(SD), Stock Repurchase(SR) and Stock Split(SS).
	}\label{tb:event}
\end{table*}

\begin{table*}[t]
	\centering
	\footnotesize
	\setlength{\tabcolsep}{2.4mm}
	\begin{tabular}{lrrrr}\toprule
		\multirow{3}{*}{\textbf{Model}} &
		\multicolumn{3}{c}{\textbf{TAE (1)}} & \multicolumn{1}{c}{\textbf{TAB (1)}}  \\
		\cmidrule(lr){2-4} \cmidrule(lr){5-5}
		& {\textbf{Win Rate}} & {\textbf{Ave. Return}} & {\textbf{Exc. Returns}}   & {\textbf{Ave. Return}} \\

        \midrule

		{\textbf{Bi-level Detection - Open} } & \textbf{54.5}$||$\textbf{34.2}\% & \textbf{1.74}\% & \$\textbf{84443} & \textbf{9.11}\%  \\
		
		\midrule

		{\textbf{Bi-level Detection - Close}} & 51.4$||$29.5\% & -0.07\% & -\$12026 & 4.50\%  \\

		\bottomrule
	\end{tabular}
	\caption{ \footnotesize  
		This table shows \textbf{Bi-level Detection}'s 1-day trading results when start transactions respectively with the \textbf{\textit{Open}} price and \textbf{\textit{Close}} price at the minute that the event is detected.
	}\label{tb:profitability}
\end{table*}

\subsection{Main Results}
Table \ref{tb:allwin1open} and \ref{tb:allwin2open} respectively shows the models' 1-day and 2-day trading results. The result of 3-day trading is consistent. Due to space limitations, we present it in \cref{sec:appendix}. 
% In these tables, \textbf{BERT-CRF} stands for the model trained and evaluated singly on the token-level task, while \textbf{Bi-level Detection} stands for the model that trained and evaluated on both article-level and token-level. 
As shown in the tables, our model outperforms all the baselines on average return and exceed return under all the settings.

\paragraph{Results of Ticker Recognizer} To evaluate the performance of ticker recognizer, we manually label tickers for 1674 news articles. The proposed ticker recognizer succeeds in 1643 of them (accuracy: $98.15\%$). Although it obtains a satisfactory performance, its imperfect recognition may slightly impair the evaluation of trading strategies, since it may point out the incorrect securities for the strategies to trade on.

\paragraph{Results of Trade-At-End} 
% Although TAE--\textit{Open} only starts the transaction less than one minute earlier than TAE--\textit{Close}, it achieves much higher profits when combined with all the event-based models. This strongly signifies the timeliness and effectiveness of the corporate event in indicating stock movement since stock prices can change rapidly in seconds when corporate events take place.

Even with the simple trading policy, our model achieves an average return of 1.74\% and an exceed return of \$84443 (844\%) in 1-day trading. 

Experiments on Vader and BERT-SST show that the sentiment of a news article can indicate the stock movement to some extend. For example, BERT-SST achieves great winning rates, and it successfully outperforms the market index by a considerable margin. However, these signals tend to results in small stock movements. Thus, sentiment-based models achieve poor average returns. 
On the other hand, event-based models obtain much higher average returns, demonstrating the superiority of corporate events over news sentiments in indicating stock movements. 
% Although the BERT-SST achieves an accuracy of 91\% in classifying positive/negative sentiments, it obtains much less average returns and exceed return in trading, which supports our argument that the news sentiments may not effectively indicate the stock movements. 

Although the Sentence model highly outperforms the sentiment-based methods, compared to our models, it is less robust to ambiguous articles, and it is more likely to miss the events that are described in several continuous sentences. By utilizing global context information and detecting events from the token level, our model identifies more trading signals and avoids more potential traps.

Bi-level Detection also achieves better performance than BERT-CRF under all the settings. The improvements mainly come from the high-level detector. By combining the global contextual information with token-level detected results, Bi-level Detection is more robust and more effective. When the low-level detector generates false alarms on some ambiguous tokens or fails to detect events that are not explicitly described, the high-level detector may point it out after analyzing the meaning of the entire article.

\paragraph{Results of Trade-At-Best} This trading policy is designed to measure the ceiling of trading signals given by each model.
Under this setting, Bi-level Detection obtains a 9.11\% average return that dramatically exceeds all the sentiment-based models, indicating how significant the stock price changes when corporate events take place. Compared to other event-based models, Bi-level Detection achieves much better performance in identifying corporate events.
% Although it is impossible to perform this policy in real-world, we can anticipate an elaborately-designed trading policy to achieve a profitability between TAE and TAB. 
% It is worth noting that when performing tradings on the Close price, the average returns of event-based methods drop greatly. This further enhances our argument that corporate events can dramatically impact stock prices in a very short time.

% It is also worth noting that even the simple Keywords model profits more than the sentiment-based models in this setting, which enhances our argument that the corporate event is a more effective indicator of stock movement.

\subsection{Event Analysis}
% We further analyze the results of trading on every single event. 
Table \ref{tb:event} shows Bi-level Detection's 1-day and 3-day trading results on each event. As shown in the table, the Dividend Increase has the smallest influence on the stock price, while the Reverse Stock Split and Clinical Trial significantly impact the stock prices. 

Reverse Stock Split and Special Dividend are the most profitable corporate event, yet they are relatively rare. Acquisition, Clinical Trial and New Contract are ubiquitous, and they also lead to significant stock movements. However, comparisons between TAE and TAB's profits indicate that these events are relatively trickier to trade. Although the ideal policy TAB achieves dramatic profits, TAE makes much fewer profits, indicating that the stocks may oscillate significantly after these events. On the other hand, Special Dividend and Reverse Stock Split are comparatively easier to trade on. 

Different events also affect the stock prices in different periods. We measure the influence by comparing TAB (1) and TAB (3)'s average return on the same event. They achieve close average returns on Acquisition, Clinical Trial, and Stock Repurchase, indicating that the stock prices do not continue to change sharply after the first day. In contrast, Stock Split impacts the stock price for a more extended period.

Thus, an ideal trading strategy should take the above factors into account. For example, it may assign different weights to each event based on their profitability and use a different policy to trade each event based on their potential price change patterns.

\subsection{Profitability in Real-world}
In this section, we discuss the possible profitability of the proposed strategy in real-world trading. Backtesting against historical data shows that the proposed strategy dramatically outperforms the market index. However, this result is based on two main assumptions.

First, we assume the cost of time in acquiring the news articles and making trading decisions is almost 0. 
Table \ref{tb:profitability} indicates the significance of timeliness, in which \textbf{Bi-level Detection - Open} starts transactions with the \textbf{\textit{Open}} price at the minute that the news article is published (e.g., price at 11:23:00), while \textbf{Bi-level Detection - Close} starts with the \textbf{\textit{Close}} price at that minute (e.g., price at 11:23:59). As shown in the table, when the model trades tens of seconds after the publish time of the news article, it greatly underperforms the market index and achieves a negative average return on each transaction. These results demonstrate that the profitability of an event-based model highly depends on how ``quick'' one can perform tradings after a piece of news is published. 

Second, we assume we can always buy/sell the desired amount of stock shares and ignore the liquidity of the stocks. When the investment scale is relatively small, this assumption doesn't have a big impact. However, as the investment scales up, the liquidity may greatly constrain the model's profitability.

% \subsection{Ablation Test}

% \begin{table}[h]
% 	\centering
% 	\footnotesize
% 	\begin{tabular}{lrr}
% 		\toprule
% 		{\textbf{Model}} & {Win Rate} & {Ave. Return}  \\
% 		\midrule
% 		Ours (TAE-\textit{Open} (1)) & 60.13\% & 2.69\%  \\
% 		~ - domain adaptation & -0.36\% & -0.46\%   \\ 
% 		~ - joint optimization & -0.87\% & -0.29\% \\
		
% 		\midrule
% 		Ours (TAE-\textit{Open} (3)) & 58.48\% & 2.79\%  \\
% 		~ - domain adaptation & -1.79\% & -0.53\%   \\
% 		~ - joint optimization & -0.32\% & -0.37\% \\
		 
% 		\bottomrule
% 	\end{tabular}
% 	\caption{
% 		\footnotesize
% 		Ablation test over different components. 
% 	}\label{tb:ablation}
% \end{table}

% We finally conduct ablation tests to examine the effectiveness of different components. When domain adaption is removed, we directly train the pre-train BERT checkpoint with joint optimization. When joint optimization is removed, we train the domain-adapted model singly on the token-level prediction. As shown in \cref{tb:ablation}, both the winning rate and average return drops greatly when joint optimization or domain adaptation is removed.

% Table \ref{tb2:eventwin} and \ref{tb3:eventprofit} respectively shows the winning rate and average return when trading on each event. As shown in the table, different event impact the stock price differently. 

\section{Related Works}
\paragraph{Text-base Stock Prediction} Existing methods usually count on textual features and sentiment analysis to forecast the stock movements \citep{HAGENAU2013685, 8848203, xie2019stock, 3155133, liu2018leveraging, mittal2012stock}. \citet{HAGENAU2013685} utilizes N-Gram, Noun-phrases, and 2-words combinations of corporate announcements to predict the stock movement. 
% \citet{mittal2012stock} predict the stock movements by analyzing Twitter sentiments based on Profile of Mood States (POMS).
The influence of financial news on the stock market is also widely explored \citep{engle1993measuring, tetlock2007giving}. In recent years, researchers resort to support vector machine and deep neural network to analyze financial news articles \citep{liu2018leveraging, xie2019stock, ding2015deep, 3155133}. \citet{8848203} combines news text (e.g., sentiment, tf-idf, and word2vec representation) with historical stock data to predict future stock prices. Event-based stock predictions are also introduced. \citet{ding2015deep} extract events from news articles, calculate event embedding, and use it to predict direction of stock moves. \citet{ben2017event} build sentiment-type trading signals with word polarities and event-type trading signals with existing information extraction platform and demonstrates the superiority of event over sentiment in making trading decisions. 

\paragraph{Event Detection} General domain event detection that aims to recognize structured schemata/frames from the text has been widely explored by data-driven supervised learning methods \citep{ahn2006stages, mitamura2015overview}. In the economic domain, however, existing approaches \citep{arendarenko2012ontology, hogenboom2013semantics, xie-etal-2013-semantic} usually exploit knowledge-based and rule-based methods, which require extensive hand-designed rules and ontology. Recent works conceptualize corporate events as sequences of text that reported company-related occurrences and introduce data-driven methods to solve financial event detection with text classification 
\citep{wiki2019financial, baseline}. \citet{wiki2019financial} 
explored a Wikipedia-based supervised method to detect the sentences that may include corporate events. \citet{baseline} propose a multi-label sequence classification model to detect specific corporate events from news articles.

\section{Conclusion}

This paper introduces an event-driven trading strategy based on corporate event detection from news articles. We introduce a bi-level event detection model that utilizes global and local information to identifies corporate events. Experiments on the presented dataset EDT demonstrate the proposed model's superiority over all the baselines. The results also signify the corporate event's timeliness and effectiveness in indicating stock movement. 

In future work, we plan to explore more on both the event detection model and trading policy. We expect to involve external knowledge and few-shot learning methods to relieve the event detection model from the data imbalance and data-scarce scenarios. On the trading policy side, we aim to explore more types of events and customize different policies for each event based on the potential price change patterns it may lead to.

% \section{Ethical Considerations}
% The presented dataset \textbf{\emph{EDT}} is introduced in details in \cref{sec:data}. EDT is collected from publicly available data sources. We mention and cite all the data sources in the main text. We are discussing with the institutional review board about the details of open-sourcing a new dataset. We expect to finish it before the author response period. 

% The data annotation is purely done by the authors of this paper. Authors are listed based on their contributions. Each annotation is checked independently by at least two human annotators to ensure its quality. We do not consider there is any potential risk with the proposed dataset.

% The proposed model does not has any potential to cause harm to vulnerable populations. We expect this model to benefit researchers in the area of financial natural language processing and social computing by pointing out a new way analyze the relationship between news articles and the stock market. We do not encourage people to invest their money in real-world based on this strategy, and we are not responsible for any potential loss in their investment.

\section*{Acknowledgments}
Han Liu is grateful for the support of National Science Foundation (BIGDATA 1840866, RI 1408910, CAREER 1841569, TRIPODS 1740735), Alfred P Sloan Fellowship, and DARPA QEDRML. We sincerely thank Polygon.io for their high-quality historical stock price data.

\bibliography{anthology,custom}
\bibliographystyle{acl_natbib}

\clearpage
\newpage
\appendix

\onecolumn
\section{Results for 3 days trading}
\label{sec:appendix}
\begin{minipage}{.5\textwidth}
As shown in table \ref{tb:allwin3open}, the trading results on 3-day trading is consistent with the results of 1-day and 2-day tradings.
\end{minipage}

\begin{table*}[h]
	\centering
	\footnotesize
	\setlength{\tabcolsep}{2.4mm}{
	\begin{tabular}{lrrrrr}\toprule
		\multirow{3}{*}{\textbf{Model}} &
		\multicolumn{3}{c}{\textbf{TAE (3)}} & \multicolumn{1}{c}{\textbf{TAB (3)}} & \\
		\cmidrule(lr){2-4} \cmidrule(lr){5-5}
		
		& {\textbf{Win Rate}} & {\textbf{Ave. Return}} & {\textbf{Exc. Returns}}   & {\textbf{Ave. Return}}  & {\textbf{Num. Trans.}}
 \\

		\midrule
		{\textbf{Vader} \citep{gilbert2014vader}}  & 55.6$||$41.9\% & 0.69\% & -\$152 & 4.11\% & 4327\\
		
		{\textbf{BERT-SST (0.995)} }   & \textbf{56.1}$||$\textbf{42.8}\% & 0.98\% & \$1643 & 8.07\% & 2378 \\
		
		{\textbf{BERT-SST (0.9)} } & 54.1$||$41.1\% & 0.73\% & \$3490 & 6.23\% & 15663 \\

		{\textbf{Sentence} \citep{baseline}}  & 54.3$||$42.6\% & 1.59\% & \$40066 & 11.11\% & 2881 \\
		
		{\textbf{BERT-CRF}}  & 51.5$||$41.2\% & 1.57\% & \$53152 & 12.94\% & 3533 \\
        
        \midrule
        
		{\textbf{Bi-level Detection} (ours)} & 52.0$||$42.0\% & \textbf{1.71}\% & \$\textbf{55339} & \textbf{13.11}\% & 3118  \\

		\bottomrule
	\end{tabular}}
	\caption{ \footnotesize  
		This table shows the 3-day trading result, in which we start each transaction at the news article's publish time and end the transaction after 3 trading day. \textbf{Win Rate} stands for the overall winning rate (rate of transactions that have a return over 0) $||$ big win rate (rate of transactions that have a return over 1\%). \textbf{Ave. Return} stands for the average return on each transaction. \textbf{Exc. Return} stands for the total excess returns over the market when starting with \$10000 and invest \$2000 to each detected trading signal. \textbf{Num. Trans.} stands for the number of transactions (valid trading signals) of each model.
	\label{tb:allwin3open}	
	}
\end{table*}

\twocolumn

\end{document}